\documentclass[10pt,twocolumn,letterpaper]{article}

\usepackage{cvpr}
\usepackage{times}
\usepackage{epsfig}
\usepackage{graphicx}
\usepackage{amsmath}
\usepackage{amssymb}
\usepackage{bbm}
\usepackage{multirow}
\usepackage{bm}
\usepackage{booktabs}
\usepackage{tabularx}
\usepackage{subcaption}

\usepackage[pagebackref=true,breaklinks=true,letterpaper=true,colorlinks,bookmarks=false]{hyperref}
\cvprfinalcopy 

\def\cvprPaperID{925} 

\ifcvprfinal\fi
\begin{document}

\title{Circle Loss: A Unified Perspective of Pair Similarity Optimization}
\newcommand*\samethanks[1][\value{footnote}]{\footnotemark[#1]}
\author{%
{\fontsize{10.5}{12.6}\selectfont Yifan Sun$^1$\thanks{Equal contribution.}, Changmao Cheng$^1$\samethanks, Yuhan Zhang$^2$\samethanks, Chi Zhang$^1$, Liang Zheng$^3$, Zhongdao Wang$^4$, Yichen Wei$^1$\thanks{Corresponding author.}}\\
{\fontsize{10.5}{12.6}\selectfont {$^1$MEGVII Technology}
{$^2$Beihang University }
{$^3$Australian National University }
{$^4$Tsinghua University}}\\
{\texttt{\small{\{peter, chengchangmao, zhangchi, weiyichen\}@megvii.com}} \hspace{0.5cm}}\\
}
\maketitle
\thispagestyle{empty}

\begin{abstract}
This paper provides a pair similarity optimization viewpoint on deep feature learning, aiming to maximize the within-class similarity $s_p$ and minimize the between-class similarity $s_n$. We find a majority of loss functions, including the triplet loss and the softmax cross-entropy loss, embed $s_n$ and $s_p$ into similarity pairs and seek to reduce $(s_n-s_p)$. Such an optimization manner is inflexible, because the penalty strength on every single similarity score is restricted to be equal. Our intuition is that if a similarity score deviates far from the optimum, it should be emphasized. 
To this end, we simply re-weight each similarity to highlight the less-optimized similarity scores. It results in a Circle loss, which is named due to its circular decision boundary. The Circle loss has a unified formula for two elemental deep feature learning paradigms, \emph {i.e.}, learning with class-level labels and pair-wise labels. Analytically, we show that the Circle loss offers a more flexible optimization approach towards a more definite convergence target, compared with the loss functions optimizing $(s_n-s_p)$. Experimentally, we demonstrate the superiority of the Circle loss on a variety of deep feature learning tasks. On face recognition, person re-identification, as well as several fine-grained image retrieval datasets, the achieved performance is on par with the state of the art.
 
\end{abstract}


\section{Introduction}\label{sec:intro}
This paper holds a similarity optimization view towards two elemental deep feature learning paradigms, \emph{i.e.}, learning from data with class-level labels and from data with pair-wise labels. The former employs a classification loss function (\emph{e.g.}, softmax cross-entropy loss~\cite{sun2014deep,Liu2016LargeMarginSL,wen2016discriminative}) to optimize the similarity between samples and weight vectors. The latter leverages a metric loss function (\emph{e.g.}, triplet loss~\cite{hoffer2015deep,schroff2015facenet}) to optimize the similarity between samples. 
In our interpretation, there is no intrinsic difference between these two learning approaches. They both seek to minimize between-class similarity $s_n$, as well as to maximize within-class similarity $s_p$.

From this viewpoint, we find that many popular loss functions (\emph{e.g.}, triplet loss~\cite{hoffer2015deep,schroff2015facenet}, softmax cross-entropy loss and its variants~\cite{sun2014deep,Liu2016LargeMarginSL,wen2016discriminative,wang2018additive,Wang_2018_CVPR,deng2019arcface}) share a similar optimization pattern. They all embed $s_n$ and $s_p$ into similarity pairs and seek to reduce $(s_n-s_p)$. In $(s_n-s_p)$, increasing $s_p$ is equivalent to reducing $s_n$. We argue that this symmetric optimization manner is prone to the following two problems.

\begin{figure}[t!]
	\centering
	\includegraphics[width=0.95\linewidth]{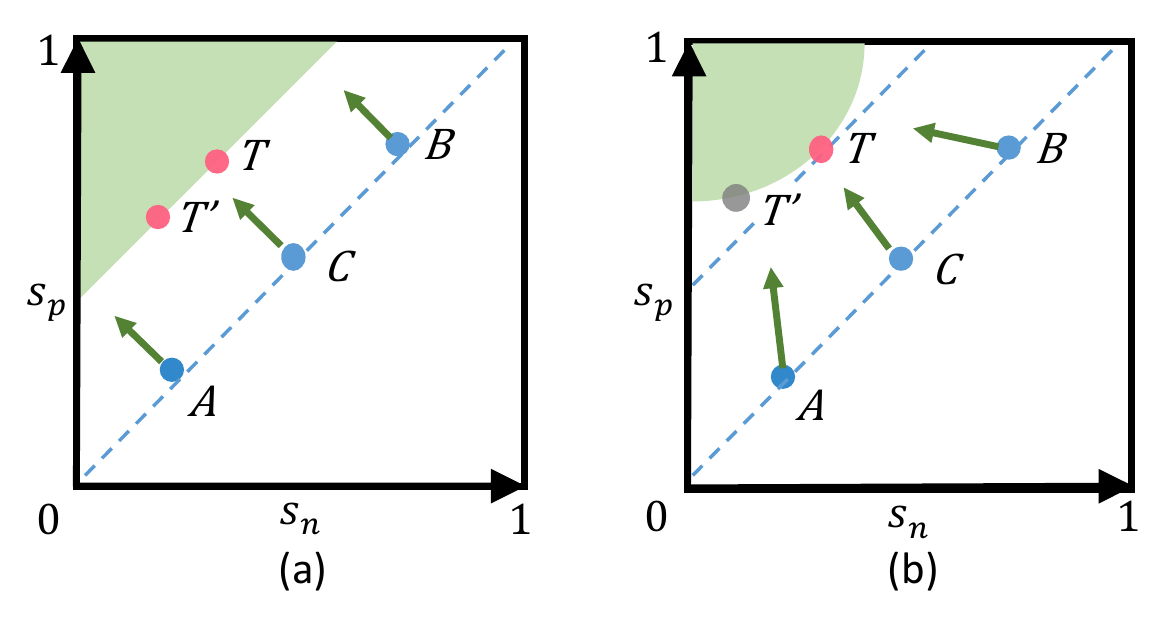}
	\caption{Comparison between the popular optimization manner of reducing $(s_n-s_p)$ and the proposed optimization manner of reducing $(\alpha_n s_n -\alpha_p s_p)$. (a) Reducing $(s_n-s_p)$ is prone to inflexible optimization ($A$, $B$ and $C$ all have equal gradients with respect to $s_n$ and $s_p$), as well as ambiguous convergence status (both $T$ and $T'$ on the decision boundary are acceptable). (b) With $(\alpha_n s_n -\alpha_p s_p)$, the Circle loss dynamically adjusts its gradients on $s_p$ and $s_n$, and thus benefits from a flexible optimization process. For $A$, it emphasizes on increasing $s_p$; for $B$, it emphasizes on reducing $s_n$. Moreover, it favors a specified point $T$ on the circular decision boundary for convergence, setting up a definite convergence target.}
	\vspace{-4mm}
	\label{fig:intro}
\end{figure}

$\bullet$ \textbf{Lack of flexibility for optimization.}  The penalty strength on $s_n$ and $s_p$ is restricted to be equal. Given the specified loss functions, the gradients with respect to $s_n$ and $s_p$ are of same amplitudes (as detailed in Section~\ref{sec:revisit}). In some corner cases, \emph{e.g.}, $s_p$ is small and $s_n$ already approaches 0 (``$A$'' in Fig.~\ref{fig:intro} (a)), it keeps on penalizing $s_n$ with a large gradient. It is inefficient and irrational. 

$\bullet$ \textbf{Ambiguous convergence status.}
Optimizing $(s_n-s_p)$ usually leads to a decision boundary of $s_p-s_n=m$ ($m$ is the margin). This decision boundary allows ambiguity (\emph{e.g.}, ``$T$'' and ``$T'$'' in Fig.~\ref{fig:intro} (a)) for convergence. For example, $T$ has $\{s_n,s_p\}=\{0.2,0.5\}$ and $T'$ has $\{s_n',s_p'\}=\{0.4,0.7\}$. They both obtain the margin $m=0.3$. However, comparing them against each other, we find the gap between $s_n'$ and $s_p$ is only $0.1$. Consequently, the ambiguous convergence compromises the separability of the feature space.

With these insights, we reach an intuition that different similarity scores should have different penalty strength. If a similarity score deviates far from the optimum, it should receive a strong penalty. Otherwise, if a similarity score already approaches the optimum, it should be optimized mildly. To this end, we first generalize $(s_n-s_p)$ into $(\alpha_n s_n - \alpha_p s_p)$, where $\alpha_n$ and $\alpha_p$ are independent weighting factors, allowing $s_n$ and $s_p$ to learn at different paces. 
We then implement $\alpha_n$ and $\alpha_p$ as linear functions \wrt $s_n$ and $s_p$ respectively, to make the learning pace adaptive to the optimization status: The farther a similarity score deviates from the optimum, the larger the weighting factor will be. Such optimization results in the decision boundary $\alpha_n s_n - \alpha_p s_p = m$, yielding a circle shape in the $(s_n,s_p)$ space, so we name the proposed loss function \emph{Circle loss}.

Being simple, Circle loss intrinsically reshapes the characteristics of the deep feature learning from the following three aspects:

\textbf{First, a unified loss function}. From the unified similarity pair optimization perspective, we propose a unified loss function for two elemental learning paradigms, \emph{learning with class-level labels and with pair-wise labels.}

\textbf{Second, flexible optimization}. During training, the gradient back-propagated to $s_n$ ($s_p$) will be amplified by $\alpha_n$ ($\alpha_p$). Those less-optimized similarity scores will have larger weighting factors and consequentially get larger gradients. As shown in Fig.~\ref{fig:intro} (b), the optimization on $A$, $B$ and $C$ are different to each other.

\textbf{Third, definite convergence status.} 
On the circular decision boundary, Circle loss favors a specified convergence status (``$T$'' in Fig.~\ref{fig:intro} (b)), as to be demonstrated in Section~\ref{sec:method_character}. Correspondingly, it sets up a definite optimization target and benefits the separability. 


The main contributions of this paper are summarized as follows:
\begin {itemize}
\item We propose Circle loss, a simple loss function for deep feature learning. By re-weighting each similarity score under supervision, Circle loss benefits the deep feature learning with flexible optimization and definite convergence target. 
\item We present Circle loss with compatibility to both class-level labels and pair-wise labels. Circle loss degenerates to triplet loss or softmax cross-entropy loss with slight modifications. 
\item We conduct extensive experiments on a variety of deep feature learning tasks, \emph{e.g.} face recognition, person re-identification, car image retrieval and so on. On all these tasks, we demonstrate the superiority of Circle loss with performance on par with the state of the art. 
\end {itemize}

\section{A Unified Perspective}\label{sec:revisit}
Deep feature learning aims to maximize the within-class similarity $s_p$, as well as to minimize the between-class similarity $s_n$. Under the cosine similarity metric, for example, we expect $s_p\rightarrow1$ and $s_n\rightarrow0$.

To this end, \textbf{learning with class-level labels} and \textbf{learning with pair-wise labels} are two elemental  paradigms. They are conventionally considered separately and significantly differ from each other \emph{w.r.t} to the loss functions. Given class-level labels, the first one basically learns to classify each training sample to its target class with a classification loss, \textit{e.g.} L2-Softmax~\cite{ranjan2017l2}, Large-margin Softmax~\cite{liu2017sphereface}, Angular Softmax~\cite{Liu2016LargeMarginSL}, NormFace~\cite{wang2017normface}, AM-Softmax~\cite{wang2018additive}, CosFace~\cite{Wang_2018_CVPR}, ArcFace~\cite{deng2019arcface}. These methods are also known as proxy-based learning, as they optimize the similarity between samples and a set of proxies representing each class. 
In contrast, given pair-wise labels, the second one directly learns pair-wise similarity (\emph{i.e.}, the similarity between samples) in the feature space and thus requires no proxies, \emph{e.g.}, constrastive loss~\cite{hadsell2006dimensionality,chopra2005learning}, triplet loss~\cite{hoffer2015deep, schroff2015facenet}, Lifted-Structure loss~\cite{oh2016deep}, N-pair loss~\cite{Sohn2016ImprovedDM}, Histogram loss~\cite{Ustinova2016LearningDE}, Angular loss~\cite{Wang2017DeepML}, Margin based loss~\cite{wu2017sampling},  Multi-Similarity loss~\cite{wang2019multi} and so on. 

This paper views both learning approaches from a unified perspective, with no preference for either proxy-based or pair-wise similarity. 
Given a single sample $x$ in the feature space, let us assume that there are $K$ within-class similarity scores and $L$ between-class similarity scores associated with $x$. We denote these similarity scores as $\{s_p^i\}\, (i=1,2,\cdots,K)$ and $\{s_n^j\}\,(j=1,2,\cdots,L)$, respectively. 

To minimize each $s_n^j$ as well as to maximize $s_p^i$, $(\forall i\in \{1,2,\cdots,K\}, \, \forall j \in \{1,2,\cdots,L\})$, we propose a unified loss function by:
\begin{equation}\label{eq:proto}
\footnotesize{
\begin{aligned}
\mathcal{L}_{uni}&=\log\Big[1+\sum_{i=1}^K\sum_{j=1}^L\exp(\gamma(s_n^j - s_p^i+m))\Big]\\
&=\log\Big[1+\sum_{j=1}^L\exp(\gamma(s_n^j+m))\sum_{i=1}^K\exp(\gamma(-s_p^i))\Big],\\
\end{aligned}
}
\end{equation} 
in which $\gamma$ is a scale factor and $m$ is a margin for better similarity separation. 

Eq.~\ref{eq:proto} is intuitive. It iterates through every similarity pair to reduce $(s_n^j-s_p^i)$. 
We note that it degenerates to triplet loss or classification loss, through slight modifications.


\begin{figure*}[t!]
	\centering
	\includegraphics[width=1\linewidth]{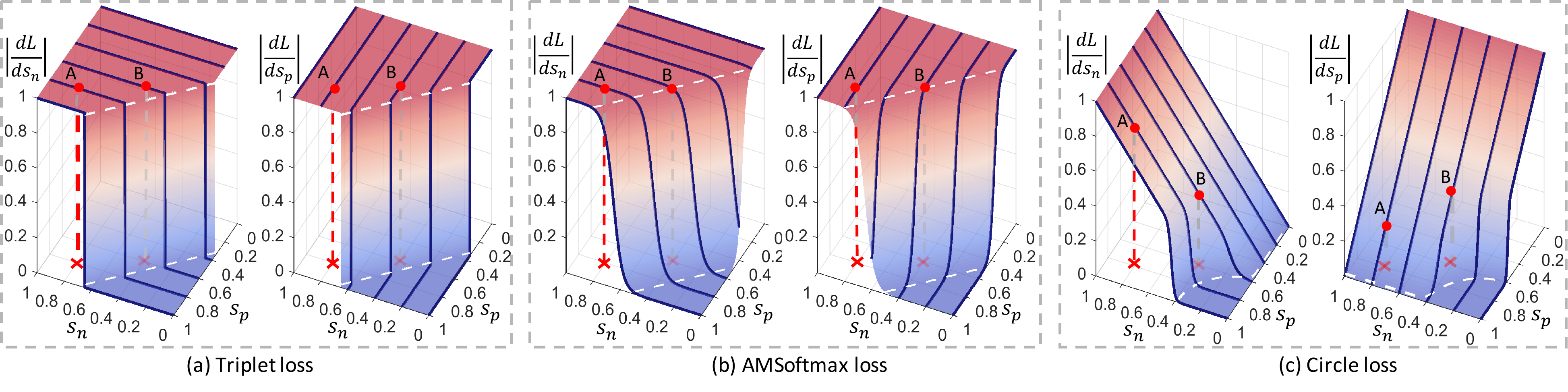}
	\caption{The gradients of the loss functions. (a) Triplet loss. (b) AM-Softmax loss. (c) The proposed Circle loss. Both triplet loss and AM-Softmax loss present the lack of flexibility for optimization. The gradients with respect to $s_p$ (left) and $s_n$ (right) are restricted to equal and undergo a sudden decrease upon convergence (the similarity pair B). For example, at $A$, the within-class similarity score $s_p$ already approaches $1$, and still incurs a large gradient. Moreover, the decision boundaries are parallel to $s_p=s_n$, which allows ambiguous convergence. In contrast, the proposed Circle loss assigns different gradients to the similarity scores, depending on their distances to the optimum. For $A$ (both $s_n$ and $s_p$ are large), Circle loss lays emphasis on optimizing $s_n$. For B, since $s_n$ significantly decreases, Circle loss reduces its gradient and thus enforces a moderated penalty. Circle loss has a circular decision boundary, and promotes accurate convergence status.}
	\vspace{-4mm}
	\label{fig:gradient}
\end{figure*}

\textbf{Given class-level labels}, we calculate the similarity scores between $x$ and weight vectors $w_i~(i=1,2,\cdots,\,N)$ ($N$ is the number of training classes) in the classification layer. Specifically, we get $(N-1)$ between-class similarity scores by: $s_n^j=w_j^\intercal x/(\|w_j\|\|x\|)$ ($w_j$ is the $j$-th non-target weight vector). Additionally, we get a single within-class similarity score (with the superscript omitted) $s_p=w_y^\intercal x/(\|w_y\|\|x\|)$. 
With these prerequisite, Eq.~\ref{eq:proto} degenerates to AM-Softmax~\cite{wang2018additive,Wang_2018_CVPR}, an important variant of Softmax loss (\emph{i.e.}, softmax cross-entropy loss):

\begin{equation}\label{eq:degenerate_softmax}
\footnotesize{
\begin{aligned}
\mathcal{L}_{am}
&=\log\Big[1+\sum_{j=1}^{N-1}\exp(\gamma (s_n^j+m))\exp(-\gamma s_p)\Big]\\
&=-\log\frac{\exp(\gamma (s_p-m))}{\exp(\gamma (s_p-m))+\sum_{j=1}^{N-1}\exp(\gamma s_n^j)}.\\
\end{aligned}
}
\end{equation} 

Moreover, with $m=0$, Eq.~\ref{eq:degenerate_softmax} further degenerates to Normface~\cite{wang2017normface}.
By replacing the cosine similarity with the inner product and setting $\gamma=1$, it finally degenerates to Softmax loss.

\textbf{Given pair-wise labels}, we calculate the similarity scores between $x$ and the other features in the mini-batch. Specifically,
$s_n^j=(x_n^j)^\intercal x/(\|x_n^j\|\|x\|)$ ($x_n^j$ is the $j$-th sample in the negative sample set $\mathcal{N}$) and $s_p^i=(x_p^i)^\intercal x/(\|x_p^i\|\|x\|)$ ($x_p^i$ is the $i$-th sample in the positive sample set $\mathcal{P}$). Correspondingly, $K=|\mathcal{P}|,\,L=|\mathcal{N}|$. Eq.~\ref{eq:proto} degenerates to triplet loss with hard mining~\cite{schroff2015facenet, hermans2017defense}:

\begin{equation}\label{eq:degenerate_triplet}
 \footnotesize{
\begin{aligned}
\mathcal{L}_{tri}&={\lim_{\gamma \to +\infty}}\frac{1}{\gamma}\mathcal{L}_{uni}\\
&={\lim_{\gamma \to +\infty}}\frac{1}{\gamma}\log\Big[1+\sum_{i=1}^{K}\sum_{j=1}^{L}\exp(\gamma (s_n^j -s_p^i+m))\Big]\\
&=\max\big[s_n^j-s_p^i+m\big]_+.\\
\end{aligned}
}
\end{equation} 

Specifically, we note that in Eq.~\ref{eq:degenerate_triplet}, the ``$\sum\exp(\cdot)$'' operation is utilized by Lifted-Structure loss~\cite{oh2016deep}, N-pair loss~\cite{Sohn2016ImprovedDM}, Multi-Similarity loss~\cite{wang2019multi} and \emph{etc.}, to conduct ``soft'' hard mining among samples. Enlarging $\gamma$ gradually reinforces the mining intensity and when $\gamma\rightarrow+\infty$, it results in the canonical hard mining in~\cite{schroff2015facenet, hermans2017defense}.

\textbf{Gradient analysis.} 
Eq.~\ref{eq:degenerate_softmax} and Eq.~\ref{eq:degenerate_triplet} show triplet loss, Softmax loss and its several variants can be interpreted as specific cases of Eq.~\ref{eq:proto}. In another word, they all optimize $(s_n-s_p)$. 
Under the toy scenario where there are only a single $s_p$ and $s_n$, we visualize the gradients of triplet loss and AM-Softmax loss in Fig.~\ref{fig:gradient} (a) and (b), from which we draw the following observations:
\begin{itemize}
\item First, before the loss reaches its decision boundary (upon which the gradients vanish), the gradients with respect to both $s_p$ and $s_n$ are the same to each other. The status $A$ has $\{s_n, s_p\}=\{0.8,0.8\}$, indicating good within-class compactness. However, $A$ still receives a large gradient with respect to $s_p$.
It leads to a lack of flexibility during optimization. 

\item Second, the gradients stay (roughly) constant before convergence and undergo a sudden decrease upon convergence. The status $B$ lies closer to the decision boundary and is better optimized, compared with $A$. However, the loss functions (both triplet loss and AM-Softmax loss) enforce an approximately equal penalty on $A$ and $B$. It is another evidence of inflexibility. 

\item Third, the decision boundaries (the white dashed lines) are parallel to $s_n-s_p=m$. Any two points (\emph{e.g.}, $T$ and $T'$ in Fig.~\ref{fig:intro}) on this boundary have an equal similarity gap of $m$, and are thus of equal difficulties to achieve. In another word, loss functions minimizing $(s_n-s_p+m)$ lay no preference on $T$ or $T'$ for convergence, and are prone to ambiguous convergence. Experimental evidence of this problem is to be accessed in Section~\ref{sec:exp_mechanism}.
\end{itemize}

These problems originate from the optimization manner of minimizing $(s_n-s_p)$, in which reducing $s_n$ is equivalent to increasing $s_p$. In the following Section~\ref{sec:circle_loss}, we will transfer such an optimization manner into a more general one to facilitate higher flexibility. 


\section{A New Loss Function} \label{sec:circle_loss}
\subsection{Self-paced Weighting}
We consider to enhance the optimization flexibility by allowing each similarity score to learn at its own pace, depending on its current optimization status. We first neglect the margin item $m$ in Eq.~\ref{eq:proto} and transfer the unified loss function into the proposed Circle loss by:
\begin{equation}\label{eq:circle}
\footnotesize{
\begin{aligned}
\mathcal{L}_{circle}&=\log\Big[1+\sum_{i=1}^K\sum_{j=1}^L\exp\big(\gamma(\alpha_n^j s_n^j - \alpha_p^i s_p^i)\big)\Big]\\
&=\log\Big[1+\sum_{j=1}^L\exp(\gamma\alpha_n^j s_n^j)\sum_{i=1}^K\exp(-\gamma\alpha_p^i s_p^i),\Big]\\
\end{aligned}
}
\end{equation} 
in which $\alpha_n^j$ and $\alpha_p^i$ are non-negative weighting factors.

Eq.~\ref{eq:circle} is derived from Eq.~\ref{eq:proto} by generalizing $(s_n^j-s_p^i)$ into $(\alpha_n^j s_n^j - \alpha_p^i s_p^i)$.
During training, the gradient with respect to $(\alpha_n^j s_n^j - \alpha_p^i s_p^i)$ is to be multiplied with $\alpha_n^j$ ($\alpha_p^i$) when back-propagated to $s_n^j$ ($s_p^i$). 
When a similarity score deviates far from its optimum (\emph{i.e.}, $O_n$ for $s_n^j$ and $O_p$ for $s_p^i$), it should get a large weighting factor so as to get effective update with large gradient.
To this end, we define $\alpha_p^i$ and $\alpha_n^j$ in a self-paced manner:

\begin{equation}\label{eq:scale}
 \footnotesize{
        \left\{
        \begin{aligned}
        \alpha_p^i=[O_p-s_p^i ]_+,\\
        \alpha_n^j=[s_n^j-O_n]_+,\\
        \end{aligned}
    \right.
    }
\end{equation}
in which $[\cdot]_+$ is the ``cut-off at zero'' operation to ensure $\alpha_p^i$ and $\alpha_n^j$ are non-negative.

\textbf{Discussions.} Re-scaling the cosine similarity under supervision is a common practice in modern classification losses~\cite{ranjan2017l2,wang2017normface,wang2018additive,Wang_2018_CVPR,Zhang2018HeatedUpSE,Zhang2019AdaCosAS}. Conventionally, all the similarity scores share an equal scale factor $\gamma$. The equal re-scaling is natural when we consider the softmax value in a classification loss function as the probability of a sample belonging to a certain class. In contrast, Circle loss multiplies each similarity score with an independent weighting factor before re-scaling. It thus gets rid of the constraint of equal re-scaling and allows more flexible optimization. Besides the benefits of better optimization, another significance of such a re-weighting (or re-scaling) strategy is involved with the underlying interpretation. Circle loss abandons the interpretation of classifying a sample to its target class with a large probability. Instead, it holds a similarity pair optimization perspective, which is compatible with two learning paradigms.

\subsection{Within-class and Between-class Margins}\label{sec:method_margin}

In loss functions optimizing $(s_n-s_p)$, adding a margin $m$ reinforces the optimization~\cite{liu2017sphereface,Liu2016LargeMarginSL,wang2018additive,Wang_2018_CVPR}. Since $s_n$ and $-s_p$ are in symmetric positions, a positive margin on $s_n$ is equivalent to  a negative margin on $s_p$. It thus only requires a single margin $m$. In Circle loss, $s_n$ and $s_p$ are in asymmetric positions. Naturally, it requires respective margins for $s_n$ and $s_p$, which is formulated by:

\begin{scriptsize}
\begin{equation}\label{eq:margin_circle}
\mathcal{L}_{circle}
=\log\big[1+\sum_{j=1}^L\exp(\gamma \alpha_n^j (s_n^j-\Delta_n))\sum_{i=1}^K\exp(-\gamma \alpha_p^i (s_p^i-\Delta_p))\big]
\end{equation} 
\end{scriptsize}
in which $\Delta_n$ and $\Delta_p$ are the between-class and within-class margins, respectively. 

Basically, Circle loss in Eq.~\ref{eq:margin_circle} expects $s_p^i>\Delta_p$ and $s_n^j<\Delta_n$. We further analyze the settings of $\Delta_n$ and $\Delta_p$ by deriving the decision boundary. For simplicity, we consider the case of binary classification, in which the decision boundary is achieved at $\alpha_n (s_n-\Delta_n)-\alpha_p (s_p- \Delta_p)=0$. 
Combined with Eq.~\ref{eq:scale}, the decision boundary is given by:
\begin{equation}\label{eq:boundary}
\footnotesize
    (s_n-\frac{O_n+\Delta_n}{2})^2 + (s_p-\frac{O_p+\Delta_p}{2})^2=C
\end{equation}
in which $C=\big((O_n-\Delta_n)^2+(O_p-\Delta_p)^2\big)/4$.

Eq.~\ref{eq:boundary} shows that the decision boundary is the arc of a circle, as shown in Fig.~\ref{fig:intro} (b). The center of the circle is at $s_n=(O_n+\Delta_n)/2, s_p=(O_p+\Delta_p)/2$, and its radius equals $\sqrt{C}$. 


There are five hyper-parameters for Circle loss, \emph{i.e.}, $O_p$, $O_n$ in Eq.~\ref{eq:scale} and $\gamma$, $\Delta_p$, $\Delta_n$ in Eq.~\ref{eq:margin_circle}. We reduce the hyper-parameters by setting $O_p=1+m$, $O_n=-m$, $\Delta_p=1-m$, and $\Delta_n=m$. 
Consequently, the decision boundary in Eq.~\ref{eq:boundary} is reduced to:

\begin{equation}\label{eq:simple_boundary}
    \begin{aligned}
    (s_n-0)^2 + (s_p-1)^2=2m^2.
    \end{aligned}
\end{equation}

With the decision boundary defined in Eq.~\ref{eq:simple_boundary}, we have another intuitive interpretation of Circle loss. It aims to optimize $s_p\rightarrow1$ and $s_n \rightarrow0$. The parameter $m$ controls the radius of the decision boundary and can be viewed as a relaxation factor. In another word, Circle loss expects $s_p^i>1-m$ and $s_n^j<m$.

Hence there are only two hyper-parameters, \emph{i.e.}, the scale factor $\gamma$ and the relaxation margin $m$.
We will experimentally analyze the impacts of $m$ and $\gamma$ in Section~\ref{sec:exp_param}.


\subsection{The Advantages of Circle Loss}\label{sec:method_character}
{The gradients} of Circle loss with respect to $s_n^j$ and $s_p^i$ are derived as follows:
\begin{small}
\begin{equation}
    \footnotesize 
    \frac{\partial \mathcal{L}_{circle}}{\partial s_n^j}
    =Z\frac{\exp\big(\gamma((s_n^j)^2-m^2)\big)}{\sum_{l=1}^L\exp\big(\gamma((s_n^l)^2-m^2)\big)}\gamma(s_n^j+m),
\end{equation}
\end{small}
and
\begin{small}
\begin{equation}\label{eq:gradient_neg}
    \footnotesize 
     \frac{\partial \mathcal{L}_{circle}}{\partial s_p^i}    =Z\frac{\exp\big(\gamma((s_p^i-1)^2-m^2)\big)}{\sum_{k=1}^K\exp\big(\gamma((s_p^k-1)^2-m^2)\big)}\gamma(s_p^i-1-m),
\end{equation}
\end{small}
in both of which
$
\footnotesize{
Z=1-\exp(-\mathcal{L}_{circle})}.
$

Under the toy scenario of binary classification (or only a single $s_n$ and $s_p$), we visualize the gradients under different settings of $m$ in Fig.~\ref{fig:gradient} (c), from which we draw the following three observations:

$\bullet~$\emph{Balanced optimization on $s_n$ and $s_p$.} We recall that the loss functions minimizing $(s_n-s_p)$ always have equal gradients on $s_p$ and $s_n$ and is inflexible. In contrast, Circle loss presents dynamic penalty strength. Among a specified similarity pair $\{s_n, s_p\}$, if $s_p$ is better optimized in comparison to $s_n$ (\emph{e.g.}, $A=\{0.8,0.8\}$ in Fig.~\ref{fig:gradient} (c)), Circle loss assigns a larger gradient to $s_n$ (and vice versa), so as to decrease $s_n$ with higher superiority. The experimental evidence of balanced optimization is to be accessed in Section~\ref{sec:exp_mechanism}.

$\bullet~$\emph{Gradually-attenuated gradients.} At the start of training, the similarity scores deviate far from the optimum and gain large gradients (\emph{e.g.}, ``$A$'' in Fig.~\ref{fig:gradient} (c)). As the training gradually approaches the convergence, the gradients on the similarity scores correspondingly decays (\emph{e.g.}, ``$B$'' in Fig.~\ref{fig:gradient} (c)), elaborating mild optimization. Experimental result in Section~\ref{sec:exp_param} shows that the learning effect is robust to various settings of $\gamma$ (in Eq.~\ref{eq:margin_circle}), which we attribute to the automatically-attenuated gradients. 

$\bullet~$\emph{A (more) definite convergence target.}
Circle loss has a circular decision boundary and favors $T$ rather than $T'$ (Fig.~\ref{fig:intro}) for convergence. It is because $T$ has the smallest gap between $s_p$ and $s_n$, compared with all the other points on the decision boundary. In another word, $T'$ has a larger gap between $s_p$ and $s_n$ and is inherently more difficult to maintain. In contrast, losses that minimize $(s_n - s_p)$ have a homogeneous decision boundary, that is, every point on the decision boundary is of the same difficulty to reach. 
Experimentally, we observe that Circle loss leads to a more concentrated similarity distribution after convergence, as to be detailed in Section \ref{sec:exp_mechanism} and Fig.~\ref{fig:scatter}.


\section{Experiments}

We comprehensively evaluate the effectiveness of Circle loss under two elemental learning approaches, \emph{i.e.}, learning with class-level labels and learning with pair-wise labels. For the former approach, we evaluate our method on face recognition (Section~\ref{sec:exp_face}) and person re-identification (Section~\ref{sec:exp_reid}) tasks. For the latter approach, we use the fine-grained image retrieval datasets (Section~\ref{sec:exp_finegrain}), which are relatively small and encourage learning with pair-wise labels. We show that Circle loss is competent under both settings. 
Section~\ref{sec:exp_param} analyzes the impact of the two hyper-parameters, \emph{i.e.}, the scale factor $\gamma$ in Eq.~\ref{eq:margin_circle} and the relaxation factor $m$ in Eq.~\ref{eq:simple_boundary}. We show that Circle loss is robust under reasonable settings. Finally, Section~\ref{sec:exp_mechanism} experimentally confirms the characteristics of Circle loss.

\subsection{Settings}

\textbf{Face recognition.}\quad We use the popular dataset MS-Celeb-1M~\cite{guo2016ms} for training. The native MS-Celeb-1M data is noisy and has a long-tailed data distribution. We clean the dirty samples and exclude few tail identities ($\le3$ images per identity). It results in $3.6M$ images and $79.9K$ identities. For evaluation, we adopt MegaFace Challenge 1 (MF1)~\cite{kemelmacher2016megaface}, IJB-C~\cite{maze2018iarpa}, LFW~\cite{LFWTech}, YTF~\cite{wolf2011face} and CFP-FP~\cite{cfp-paper} datasets and the official evaluation protocols are used. We also polish the probe set and 1M distractors on MF1 for more reliable evaluation, following~\cite{deng2019arcface}. 
For data pre-processing, we resize the aligned face images to $112\times112$ and linearly normalize the pixel values of RGB images to $[-1,1]$~\cite{wen2016discriminative,liu2017sphereface,Wang_2018_CVPR}. We only augment the training samples by random horizontal flip. We choose the popular residual networks~\cite{he2016deep} as our backbones.
All the models are trained with 182k iterations. The learning rate is started with 0.1 and reduced by 10$\times$ at 50\%, 70\% and 90\% of total iterations respectively. The default hyper-parameters of our method are $\gamma=256$ and $m=0.25$ if not specified. 
For all the model inference, we extract the 512-D feature embeddings and use cosine distance as the metric.

\textbf{Person re-identification.}\quad 
Person re-identification (re-ID) aims to spot the appearance of the same person in different observations. 
We evaluate our method on two popular datasets, \emph{i.e.}, Market-1501~\cite{Zheng_2015_ICCVmarket} and MSMT17~\cite{Wei_2018_CVPRMSMT17}. Market-1501 contains 1,501 identities, 12,936 training images and 19,732 gallery images captured with 6 cameras. MSMT17 contains 4,101 identities, 126,411 images captured with 15 cameras and presents a long-tailed sample distribution. We adopt two network structures, \emph{i.e.} a global feature learning model backboned on ResNet50 and a part-feature model named MGN~\cite{Wang_2018MGN}. We use MGN with consideration of its competitive performance and relatively concise structure. The original MGN uses a Sofmax loss on each part feature branch for training. Our implementation concatenates all the part features into a single feature vector for simplicity. For Circle loss, we set $\gamma=128$ and $m=0.25$.

\textbf{Fine-grained image retrieval.}\quad We use three datasets for evaluation on fine-grained image retrieval, \textit{i.e.} CUB-200-2011~\cite{WahCUB_200_2011}, Cars196~\cite{krause20133d} and Stanford Online Products~\cite{oh2016deep}. 
CARS-196 contains $16,183$ images which belong to $196$ class of cars. The first $98$ classes are used for training and the last $98$ classes are used for testing. CUB-200-2010 has $200$ different class of birds. We use the first $100$ class with $5,864$ images for training and the last $100$ class with $5,924$ images for testing. SOP is a large dataset that consists of $120,053$ images belonging to $22,634$ classes of online products. The training set contains $11,318$ class includes $59,551$ images and the rest $11,316$ class includes $60,499$ images are for testing.
The experimental setup follows~\cite{oh2016deep}. We use BN-Inception~\cite{ioffe2015batch} as the backbone to learn 512-D embeddings. We adopt P-K sampling trategy~\cite{hermans2017defense} to construct mini-batch with $P=16$ and $K=5$. 
For Circle loss, we set $\gamma=80$ and $m=0.4$.

\begin{table}[t]
    \small
    \centering
    \caption{Face identification and verification results on MFC1 dataset. ``Rank 1'' denotes rank-1 identification accuracy. ``Veri.'' denotes verification TAR (True Accepted Rate) at 1e-6 FAR (False Accepted Rate) with $1M$ distractors. ``R34'' and ``R100'' denote using ResNet34 and ResNet100 backbones, respectively.}
    \label{tab:mf1}
    \begin{tabularx}{\linewidth}{Xcccc}
    \toprule
     \multirow{2}{*}{Loss function} & \multicolumn{2}{c}{Rank 1 (\%)} &
     \multicolumn{2}{c}{Veri. (\%)}\\
    \cmidrule(l{2pt}r{2pt}){2-3} \cmidrule(l{2pt}r{2pt}){4-5}
     & R34 & R100 &R34 & R100  \\
     \midrule
     
    Softmax &92.36 &95.04& 92.72 &95.16 \\
    NormFace~\cite{wang2017normface} &92.62  &95.27 &92.91 &95.37 \\
    AM-Softmax~\cite{wang2018additive,Wang_2018_CVPR} &97.54 &98.31 & 97.64 &98.55 \\
    ArcFace~\cite{deng2019arcface} &97.68 & 98.36 &97.70 &98.58 \\
    CircleLoss (ours) &\textbf{97.81} &\textbf{98.50}&\textbf{98.12} &\textbf{98.73} \\
    \bottomrule
    \end{tabularx}
\end{table}
\begin{table}[t]
    \small
    \centering
    \caption{Face verification accuracy (\%) on LFW, YTF and CFP-FP with ResNet34 backbone.}
    \label{tab:face-verif}
    \begin{tabularx}{\linewidth}{lccc}
    \toprule
     Loss function & LFW~\cite{LFWTech}& YTF~\cite{wolf2011face} & CFP-FP~\cite{cfp-paper} \\
     \midrule
    Softmax &99.18 & 96.19& 95.01\\
    NormFace~\cite{wang2017normface} & 99.25 & 96.03 & 95.34\\
    AM-Softmax~\cite{wang2018additive,Wang_2018_CVPR} & 99.63 & 96.31 & 95.78 \\
    ArcFace~\cite{deng2019arcface} & 99.68 & 96.34 & 95.84\\
    CircleLoss(ours) & \textbf{99.73} & \textbf{96.38} & \textbf{96.02} \\
    \bottomrule
    \end{tabularx}
\end{table}

\subsection{Face Recognition}\label{sec:exp_face}

For face recognition task, we compare Circle loss against several popular classification loss functions, \emph{i.e.}, vanilla Softmax, NormFace~\cite{wang2017normface}, AM-Softmax~\cite{wang2018additive} (or CosFace~\cite{Wang_2018_CVPR}), ArcFace~\cite{deng2019arcface}. Following the original papers~\cite{wang2018additive, deng2019arcface}, we set $\gamma=64,m=0.35$ for AM-Softmax and $\gamma=64, m=0.5$ for ArcFace. 

\begin{table}[t]
    \small
    \centering
    \caption{Comparison of TARs on the IJB-C 1:1 verification task.}
    \label{tab:ijb-c}
    \begin{tabularx}{\linewidth}{Xccc}
    \toprule
     \multirow{2}{*}{Loss function} & \multicolumn{3}{c}{TAR@FAR (\%)} \\
     \cmidrule{2-4}
     & 1e-3 & 1e-4 & 1e-5 \\
     \midrule
    ResNet34, AM-Softmax~\cite{wang2018additive,Wang_2018_CVPR} & 95.87 & 92.14 & 81.86\\
    ResNet34, ArcFace~\cite{deng2019arcface} 	&95.94 & 92.28 & 84.23\\
    ResNet34, CircleLoss(ours) & \textbf{96.04} & \textbf{93.44} & \textbf{86.78} \\
    \hline
    ResNet100, AM-Softmax~\cite{wang2018additive,Wang_2018_CVPR} & 95.93 & 93.19 & 88.87\\
    ResNet100, ArcFace~\cite{deng2019arcface} 	&96.01 & 93.25 & 89.10\\
    ResNet100, CircleLoss(ours) & \textbf{96.29} & \textbf{93.95} & \textbf{89.60} \\
    \bottomrule
    \end{tabularx}
\end{table}

We report the identification and verification results on MegaFace Challenge 1 dataset (MFC1) in Table~\ref{tab:mf1}. Circle loss marginally outperforms the counterparts under different backbones. For example, with ResNet34 as the backbone, Circle loss surpasses the most competitive one (ArcFace) by +0.13\% at rank-1 accuracy. With ResNet100 as the backbone, while ArcFace achieves a high rank-1 accuracy of 98.36\%, Circle loss still outperforms it by +0.14\%. The same observations also hold for the verification metric.

Table~\ref{tab:face-verif} summarizes face verification results on LFW~\cite{LFWTech}, YTF~\cite{wolf2011face} and CFP-FP~\cite{cfp-paper}.
We note that performance on these datasets is already near saturation. Specifically, ArcFace is higher than AM-Softmax by +0.05\%, +0.03\%, +0.07\% on three datasets, respectively. Circle loss remains the best one, surpassing ArcFace by +0.05\%, +0.06\% and +0.18\%, respectively.

We further compare Circle loss with AM-Softmax and ArcFace on IJB-C 1:1 verification task in Table~\ref{tab:ijb-c}. Under both ResNet34 and ResNet100 backbones, Circle loss presents considerable superiority. For example, with ResNet34, Circle loss significantly surpasses ArcFace by +1.16\% and +2.55\% on ``TAR@FAR=1e-4'' and ``TAR@FAR=1e-5'', respectively.

\begin{table}[t]
    \small
    \centering
    \caption{Evaluation of Circle loss on re-ID task. We report R-1 accuracy (\%) and mAP (\%). }
    \label{tab:person-reid}
    \begin{tabular}{lcccc}
    \toprule
     \multirow{2}{*}{Method} & \multicolumn{2}{c}{Market-1501} & \multicolumn{2}{c}{MSMT17}\\
     \cmidrule{2-5}
     & R-1 & mAP & R-1& mAP \\
     \midrule
     PCB~\cite{Sun_2018_ECCVPCB} (Softmax)&93.8&81.6&68.2&40.4\\
     MGN~\cite{Wang_2018MGN} (Softmax+Triplet) &95.7&86.9&-&-\\
     JDGL~\cite{Zheng_2019_CVPRJDGL} &94.8&86.0&\textbf{77.2}&\textbf{52.3}\\

    ResNet50 + AM-Softmax &92.4&83.8&75.6&49.3\\
    ResNet50 + CircleLoss(ours) &94.2& 84.9 &76.3&50.2\\
    MGN + AM-Softmax &95.3&86.6&76.5&51.8\\
    MGN + CircleLoss(ours) &\textbf{96.1}&\textbf{87.4}&{76.9}&{52.1}\\
    \bottomrule
    \end{tabular}
\end{table}

\begin{table*}[t]
    \small
    \centering
    \caption{Comparison of R@K(\%) on three fine-grained image retrieval datasets. Superscript denotes embedding size.}
    \label{tab:cub-cars}
    \begin{tabularx}{\textwidth}{Xcccccccccccccc}
    \toprule
     \multirow{2}{*}{Loss function} & \multicolumn{4}{c}{CUB-200-2011~\cite{WahCUB_200_2011}} && \multicolumn{4}{c}{Cars196~\cite{krause20133d}} && \multicolumn{4}{c}{Stanford Online Products~\cite{oh2016deep}}\\
     \cmidrule{2-5} \cmidrule{7-10} \cmidrule{12-15}
  & R@1 & R@2 & R@4 & R@8 &  & R@1 & R@2 & R@4 & R@8 &  & R@1 &R@10& R@$10^2$ & R@$10^3$\\
     \midrule
     
    LiftedStruct$^{64}$~\cite{oh2016deep}      &43.6 &56.6&68.6&79.6  &&53.0&65.7&76.0&84.3  &&62.5&80.8&91.9&97.4 \\
    HDC$^{384}$~\cite{Song_2017_CVPRHDC}              &53.6 &65.7&77.0&85.6  &&73.7&83.2&89.5&93.8  &&69.5&84.4&92.8&97.7\\
    HTL$^{512}$~\cite{Ge_2018_ECCVHTL}               &57.1 &68.8&78.7&86.5  &&81.4&88.0&92.7&95.7  &&74.8&88.3&94.8&98.4\\
    ABIER$^{512}$~\cite{ABIER}   &57.5 &71.5&79.8&87.4  &&82.0&89.0&93.2&96.1  &&74.2&86.9&94.0&97.8\\
    ABE$^{512}$~\cite{Kim_2018_ECCVABE}    &60.6 &71.5&79.8&87.4  &&\textbf{85.2}&\textbf{90.5}&94.0&96.1  &&76.3&88.4&94.8&98.2\\ 
    Multi-Simi$^{512}$~\cite{wang2019multi} & 65.7 & 77.0 & \textbf{86.3} & 91.2 &&   {84.1} & {90.4} & 94.0 & 96.5 && 78.2 & 90.5 & 96.0 & \textbf{98.7}\\
    CircleLoss$^{512}$                 & \textbf{66.7} & \textbf{77.4} & 86.2 & \textbf{91.2} &&   83.4 & 89.8 & \textbf{94.1} & \textbf{96.5} && \textbf{78.3} & \textbf{90.5} & \textbf{96.1} & 98.6 \\

    \bottomrule
    \end{tabularx}
\end{table*}

\subsection{Person Re-identification}\label{sec:exp_reid}

We evaluate Circle loss on re-ID task in Table~\ref{tab:person-reid}. MGN~\cite{Wang_2018MGN} is one of the state-of-the-art methods and is featured for learning multi-granularity part-level features. Originally, it uses both Softmax loss and triplet loss to facilitate joint optimization. Our implementation of ``MGN (ResNet50) + AM-Softmax'' and ``MGN (ResNet50)+ Circle loss'' only use a single loss function for simplicity. 

We make three observations from Table~\ref{tab:person-reid}. First, we find that Circle loss can achieve competitive re-ID accuracy against state of the art. We note that ``JDGL'' is slightly higher than ``MGN + Circle loss'' on MSMT17~\cite{Wei_2018_CVPRMSMT17}. JDGL~\cite{Zheng_2019_CVPRJDGL} uses a generative model to augment the training data, and significantly improves re-ID over the long-tailed dataset. Second, comparing Circle loss with AM-Softmax, we observe the superiority of Circle loss, which is consistent with the experimental results on the face recognition task. Third, comparing ``ResNet50 + Circle loss'' against ``MGN + Circle loss'', we find that part-level features bring incremental improvement to Circle loss. It implies that Circle loss is compatible with the part-model specially designed for re-ID.

\subsection{Fine-grained Image Retrieval}\label{sec:exp_finegrain}
\vspace{0.5em}
We evaluate the compatibility of Circle loss to pair-wise labeled data on three fine-grained image retrieval datasets, \emph{i.e.}, CUB-200-2011, Cars196, and Standford Online Products. On these datasets, majority methods~\cite{oh2016deep,Song_2017_CVPRHDC,Ge_2018_ECCVHTL,ABIER,Kim_2018_ECCVABE,wang2019multi} adopt the encouraged setting of learning with pair-wise labels. We compare Circle loss against these state-of-the-art methods in Table~\ref{tab:cub-cars}. We observe that Circle loss achieves competitive performance, on all of the three datasets. Among the competing methods, LiftedStruct~\cite{oh2016deep} and Multi-Simi~\cite{wang2019multi} are specially designed with elaborate hard mining strategies for learning with pair-wise labels. HDC~\cite{Song_2017_CVPRHDC}, ABIER~\cite{ABIER} and ABE~\cite{Kim_2018_ECCVABE} benefit from model ensemble. In contrast, the proposed Circle loss achieves performance on par with the state of the art, without any bells and whistles. 

\begin{figure}[t]
\centering
  \includegraphics[width=\linewidth]{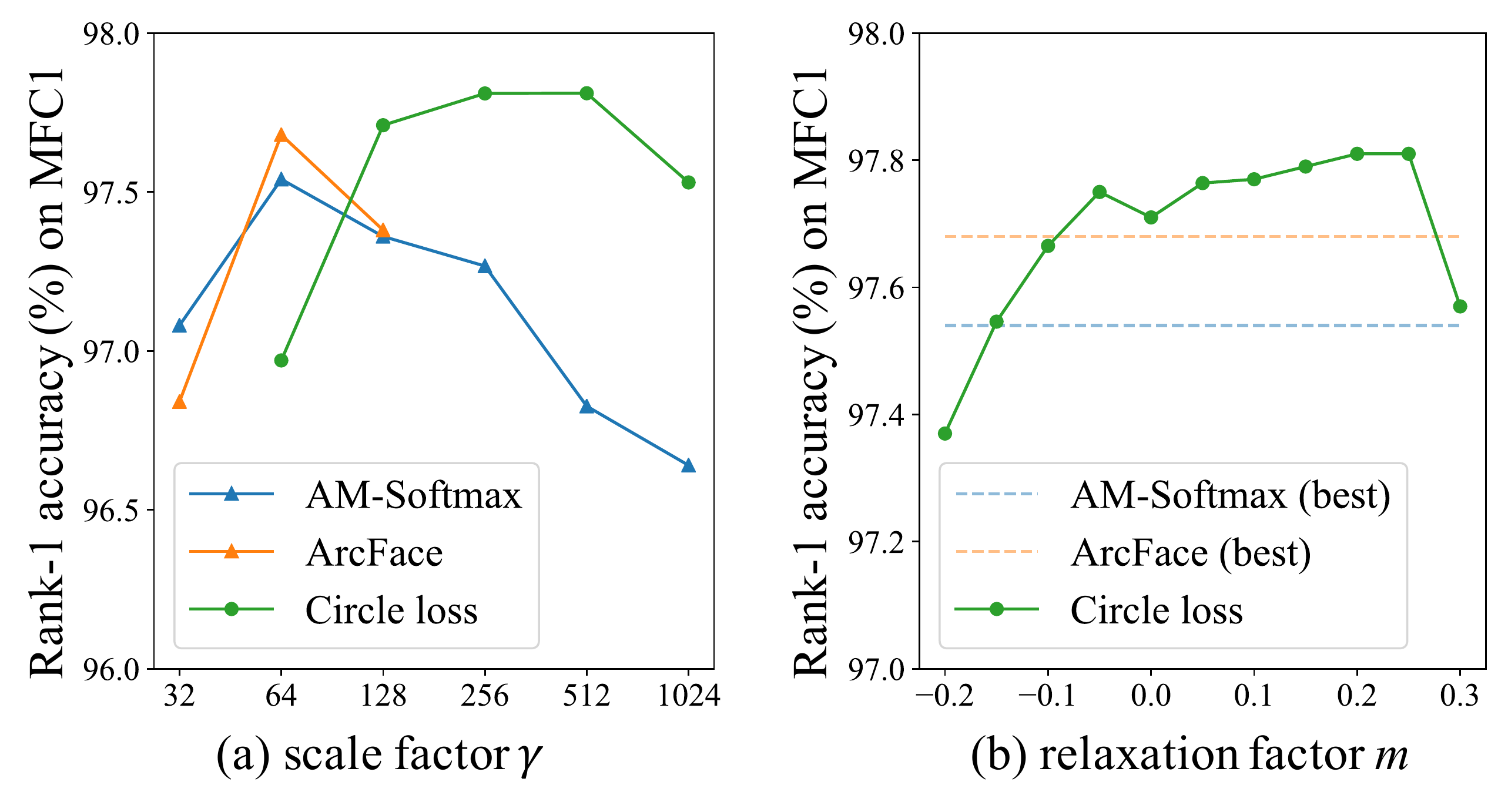}  
  \caption{Impact of two hyper-parameters. In (a), Circle loss presents high robustness on various settings of scale factor $\gamma$. In (b), Circle loss surpasses the best performance of both AM-Softmax and ArcFace within a large range of relaxation factor $m$. }
\label{fig:params}
\end{figure}

\subsection{Impact of the Hyper-parameters}\label{sec:exp_param}
We analyze the impact of two hyper-parameters, \emph{i.e.}, the scale factor $\gamma$ in Eq.~\ref{eq:margin_circle} and the relaxation factor $m$ in Eq.~\ref{eq:simple_boundary} on face recognition tasks.

\begin{figure}[t]
    \centering
    \includegraphics[width=\linewidth]{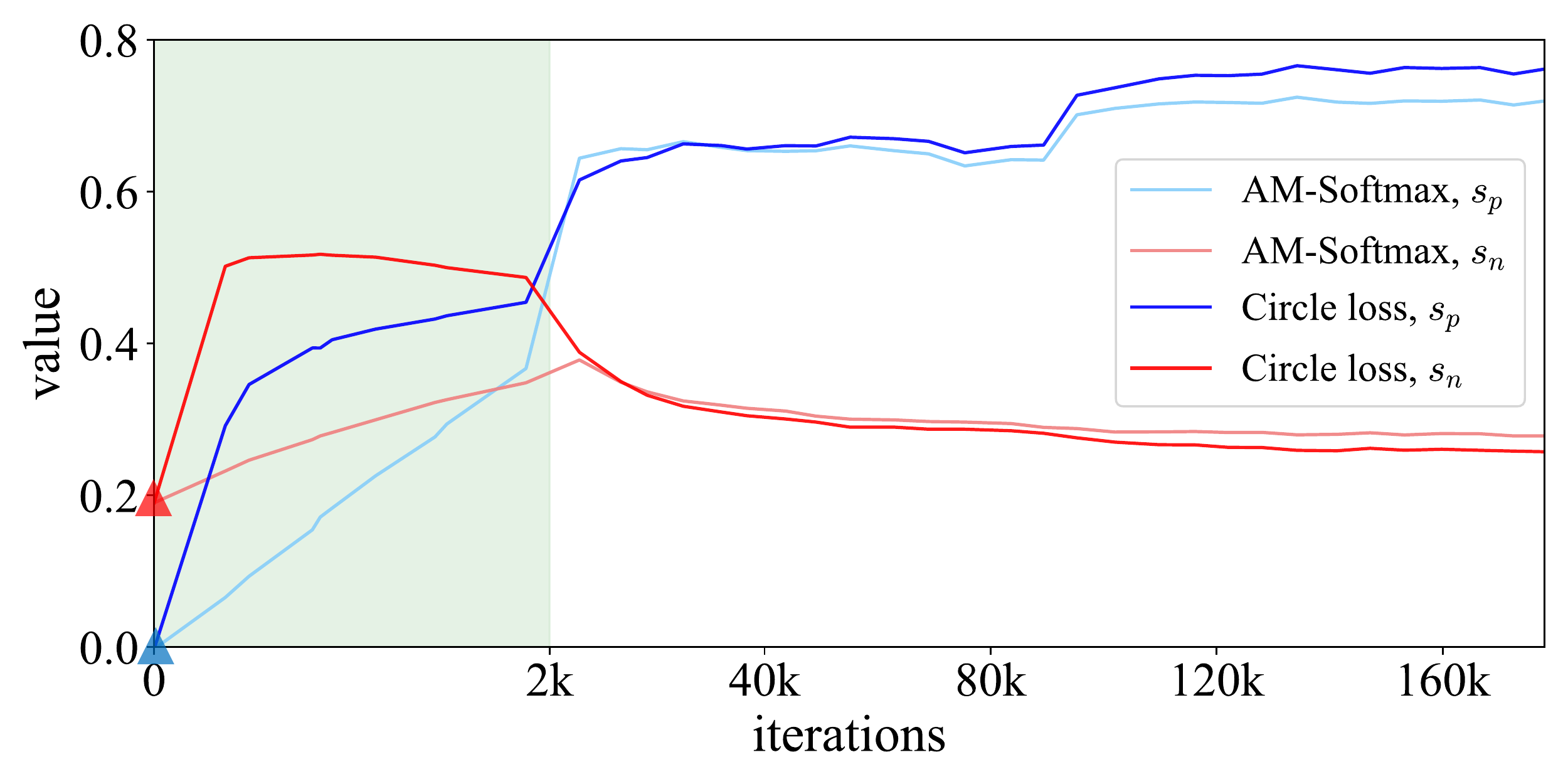}
    \caption{The change of $s_p$ and $s_n$ values during training. We linearly lengthen the curves within the first 2$k$ iterations to highlight the initial training process (in the \textcolor{green}{green} zone). During the early training stage, Circle loss rapidly increases $s_p$, because $s_p$ deviates far from the optimum at the initialization and thus attracts higher optimization priority.}
    \label{fig:logit-train}
\end{figure}

\begin{figure*}[ht]
\centering
  \includegraphics[width=0.9\linewidth]{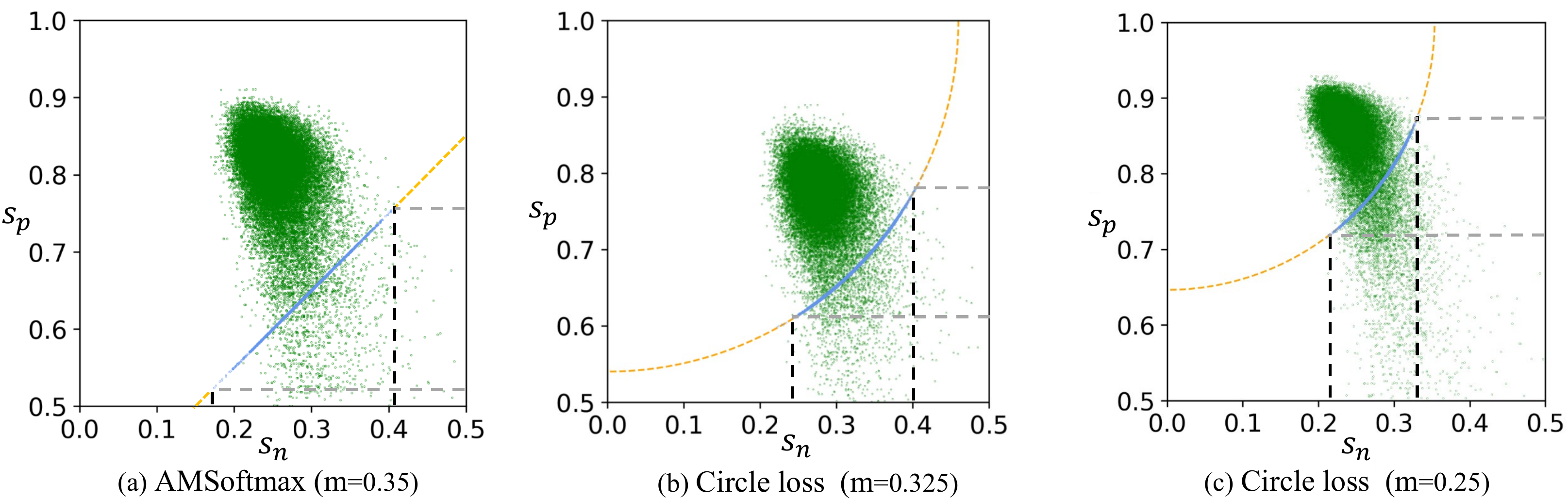}
\caption{Visualization of the similarity distribution after convergence. The \textcolor{blue}{blue} dots mark the similarity pairs crossing the decision boundary during the whole training process. The \textcolor{green}{green} dots mark the similarity pairs after convergence. (a) AM-Softmax seeks to minimize $(s_n-s_p)$. During training, the similarity pairs cross the decision boundary through a wide passage. After convergence, the similarity pairs scatter in a relatively large region in the $(s_n, s_p)$ space. In (b) and (c), Circle loss has a circular decision boundary. The similarity pairs cross the decision boundary through a narrow passage and gather into a relatively concentrated region. }
\label{fig:scatter}
\end{figure*}

\textbf{The scale factor $\gamma$} determines the largest scale of each similarity score. The concept of the scale factor is critical in a lot of variants of Softmax loss. We experimentally evaluate its impact on Circle loss and make a comparison with several other loss functions involving scale factors. We vary $\gamma$ from $32$ to $1024$ for both AM-Softmax and Circle loss. For ArcFace, we only set $\gamma$ to 32, 64 and 128, as it becomes unstable with larger $\gamma$ in our implementation. The results are visualized in Fig.~\ref{fig:params}. Compared with AM-Softmax and ArcFace, Circle loss exhibits high robustness on $\gamma$. The main reason for the robustness of Circle loss on $\gamma$ is the automatic attenuation of gradients. As the similarity scores approach the optimum during training, the weighting factors gradually decrease. Consequentially, the gradients automatically decay, leading to a moderated optimization.

\textbf{The relaxation factor $m$} determines the radius of the circular decision boundary. We vary $m$ from $-0.2$ to $0.3$ (with $0.05$ as the interval) and visualize the results in Fig.~\ref{fig:params} (b). It is observed that under all the settings from $-0.05$ to $0.25$, Circle loss surpasses the best performance of Arcface, as well as AM-Softmax, presenting a considerable degree of robustness.

\subsection{Investigation of the Characteristics}\label{sec:exp_mechanism}

\textbf{Analysis of the optimization process.}\quad
To intuitively understand the learning process, we show the change of $s_n$ and $s_p$ during the whole training process in Fig.~\ref{fig:logit-train}, from which we draw two observations:

First, at the initialization, all the $s_n$ and $s_p$ scores are small. It is because randomized features are prone to be far away from each other in the high dimensional feature space~\cite{Zhang2019AdaCosAS,helanqing_dissection}. Correspondingly, $s_p$ get significantly larger weights (compared with $s_n$), and the optimization on $s_p$ dominates the training, incurring a fast increase in similarity values in Fig.~\ref{fig:logit-train}. This phenomenon evidences that Circle loss maintains a flexible and balanced optimization.
 
Second, at the end of the training, Circle loss achieves both better within-class compactness and between-class discrepancy (on the training set), compared with AM-Softmax. Because Circle loss achieves higher performance on the testing set, we believe that it indicates better optimization.

\textbf{Analysis of the convergence.}\quad
We analyze the convergence status of Circle loss in Fig.~\ref{fig:scatter}.
We investigate two issues: how the similarity pairs consisted of $s_n$ and $s_p$ cross the decision boundary during training and how they are distributed in the $(s_n, s_p)$ space after convergence. The results are shown in Fig.~\ref{fig:scatter}. In Fig.~\ref{fig:scatter} (a), AM-Softmax loss adopts the optimal setting of $m=0.35$. In Fig.~\ref{fig:scatter} (b), Circle loss adopts a compromised setting of $m=0.325$. The decision boundaries of (a) and (b) are tangent to each other, allowing an intuitive comparison. In Fig.~\ref{fig:scatter} (c), Circle loss adopts its optimal setting of $m=0.25$. Comparing Fig.~\ref{fig:scatter} (b) and (c) against Fig.~\ref{fig:scatter} (a), we find that Circle loss presents a relatively narrower passage on the decision boundary, as well as a more concentrated distribution for convergence (especially when $m=0.25$). It indicates that Circle loss facilitates more consistent convergence for all the similarity pairs, compared with AM-Softmax loss. 
This phenomenon confirms that Circle loss has a more definite convergence target, which promotes the separability in the feature space.

\section{Conclusion}
This paper provides two insights into the optimization process for deep feature learning. First, a majority of loss functions, including the triplet loss and popular classification losses, conduct optimization by embedding the between-class and within-class similarity into similarity pairs. Second, within a similarity pair under supervision, each similarity score favors different penalty strength, depending on its distance to the optimum. These insights result in Circle loss, which allows the similarity scores to learn at different paces. The Circle loss benefits deep feature learning with high flexibility in optimization and a more definite convergence target. It has a unified formula for two elemental learning approaches, \emph{i.e.}, learning with class-level labels and learning with pair-wise labels. On a variety of deep feature learning tasks, \emph{e.g.}, face recognition, person re-identification, and fine-grained image retrieval, the Circle loss achieves performance on par with the state of the art. 


{\small
\bibliographystyle{ieee}
\bibliography{egbib}
}

 
\vspace{0.5em}

\end{document}